\documentclass[11pt]{article}
\usepackage[margin=1in]{geometry}
\usepackage[utf8]{inputenc}
\usepackage[T1]{fontenc}
\usepackage{amsmath,amssymb,amsthm}
\usepackage{graphicx}
\usepackage{booktabs}
\usepackage{hyperref}
\usepackage{array}
\usepackage{multirow}
\usepackage{pgfplots}
\pgfplotsset{compat=1.18}
\newtheorem{proposition}{Proposition}

\title{\textbf{Attention-Mass Condensation for Sparse Decoding}\\[0.4em]
\large Margin Stability, Query-Dependent Retrieval, and Operating Limits}
\author{Jorge L. Ruiz Williams}
\date{}

\begin{document}
\maketitle
\begin{abstract}
Attention-mass concentration creates an opportunity for sparse decoding, but retained mass alone does not guarantee a stable greedy decision: retrieval error, omitted value directions, and recursive decoding all matter. We formalize this distinction with an exact omitted-mass identity and a sufficient downstream margin condition, then characterize a query-dependent mean-pooled block selector. On Qwen2-0.5B, a paired fresh-selection sweep covers supports of 97--769 positions, contexts of 2K--16K, and five prefixes per context. The primary exact-match result is that none of 60 runs remains identical to dense decoding through 128 tokens. Distributional quality is distinct: for supports of at least 193, seven of nine context-support conditions have median teacher-forced continuation perplexity changes within 5\% of dense, but prompt-level ranges include severe 16K outliers. All seven runs with teacher-forced match below 70\% have perplexity increases above 100\%; these observations come from two prefixes and suggest a warning regime, not a general threshold. The measured perplexity is teacher-forced on the dense model's own continuation, not the sparse model's free-running output. Separate retrieval and attention-mass probes illustrate why captured mass alone is not a retrieval or decision guarantee. Isolated operator timings do not establish matched-quality acceleration or end-to-end serving speed.
\end{abstract}

\section{Introduction}
Long-context inference has two different attention costs. Prefill evaluates interactions across the prompt and scales as $O(N^2d)$ per head. During autoregressive decode, each new query interacts with the historical KV cache, giving $O(Nd)$ attention work and memory traffic per head. Exact kernels such as FlashAttention reduce I/O overhead, but they still evaluate dense attention \cite{dao2022flashattention}. As contexts grow, avoiding unnecessary KV reads becomes increasingly attractive.

The basic observation behind sparse attention is not new: trained transformers often place most of their attention mass on a relatively small set of positions. That alone is not enough. A small omitted weight can still matter if its value contribution is important, a coarse selector can miss the right distant key, and a tiny local error can eventually flip an autoregressive decision.

We use the term \textbf{Condensation Principle} for one specific claim: \emph{when a trained model's attention is sufficiently concentrated on a small query-dependent subset, and the contribution of everything omitted remains inside the model's downstream decision margin, sparse attention can preserve the dense greedy decision}. ``Condensation'' refers only to attention mass concentrating onto a small support. It is not a claim that attention is universally sparse, that a fixed support size is always sufficient, or that a particular retrieval rule will recover the right support.

This paper asks where that principle holds and where it stops holding for a concrete low-budget selector. Our reference operating point keeps the first token, a 64-token local window, and two distant 16-token blocks chosen by scoring the current query against mean-pooled post-RoPE key blocks, for 97 positions per head. We vary this selector's support budget; other policies are specified as budget-matched alternatives but are not evaluated in the clean survival sweep.

The primary result is exact-match instability: all 60 fresh-selection trajectories diverge from dense decoding within 43 sparse decisions. This is separate from distributional quality. For support $S\ge193$, seven of nine context-support conditions have median continuation perplexity changes within 5\% of dense, while the per-prompt ranges include severe 16K failures. In this sample, every run with teacher-forced match below 70\% has $\Delta$PPL above 100\%, but those runs come from only two 16K prefixes. We treat this as a descriptive warning regime to test, not a general threshold. The 97--769-position curve characterizes a model- and prompt-dependent operating envelope.

We also report isolated timing measurements because they show why the idea remains computationally interesting. Candidate-given sparse attention can be much cheaper than dense attention, and fresh query-dependent selection can already beat dense single-query attention at long contexts. But these timing measurements are not end-to-end matched-quality results. The paper therefore separates two questions throughout: \emph{how much computation can be avoided?} and \emph{what quality is lost when we do so?}

\paragraph{Contributions.}
The paper makes four connected contributions. First, we formulate attention-mass condensation as a conditional greedy-stability principle, using an exact omitted-mass identity and a sufficient margin condition. Second, we specify a practical selector built from anchor and local retention with query-dependent mean-pooled block retrieval, and define budget-matched alternatives without claiming that any is optimal. Third, we characterize the tested KV-group selector's support operating curve and connect it to separate retrieval and mass-recovery diagnostics, distinguishing captured mass from useful retrieval. Fourth, we report isolated attention timings separately from quality and state what those measurements do not establish.

\section{Decision Stability}
For a fixed query, let $\mathcal{C}$ be the retained positions and $\mathcal{D}$ the omitted positions. Let
\[
\varepsilon = \sum_{j\in\mathcal{D}} A^{\mathrm{full}}_{ij}
\]
be the full-attention mass assigned to the omitted set. Define $o_C$ and $o_D$ as the value averages normalized within $\mathcal{C}$ and $\mathcal{D}$. Then
\begin{equation}
o^{\mathrm{full}}=(1-\varepsilon)o_C+\varepsilon o_D,
\end{equation}
and sparse attention over $\mathcal{C}$ gives $o^{\mathrm{sparse}}=o_C$. Therefore
\begin{equation}
\Delta o = o^{\mathrm{sparse}}-o^{\mathrm{full}} = \varepsilon(o_C-o_D),
\end{equation}
so if $V_{\max}=\max_j\|v_j\|_2$,
\begin{equation}
\|\Delta o\|_2\le 2\varepsilon V_{\max}.
\label{eq:single_layer_bound}
\end{equation}
This identity is exact in real arithmetic. It does not say that omitted values are individually negligible, only that their aggregate effect is scaled by the omitted attention mass.

Across a deep network, earlier perturbations change later hidden states. For a single forward evaluation with a fixed historical cache, consider the usual hybrid sequence in which attention layers are replaced by sparse attention one at a time. For query position $i$, define the concatenated multi-head perturbation bound
\[
E_l \equiv \left(\sum_{h=1}^{H_l}\left(2\varepsilon_{l,h}V_{\max,l,h}\right)^2\right)^{1/2},
\]
where $\varepsilon_{l,h}$ is the omitted attention mass and $V_{\max,l,h}$ bounds the value-vector norm for head $h$ at layer $l$. This follows from the per-head omitted-mass bound and the Euclidean norm of concatenated heads. Let $\kappa_l$ bound the downstream gain from the concatenated attention output at layer $l$ to final logits, including the output projection. Then triangle inequality gives
\begin{equation}
\|\Delta \ell\|_\infty \le \sum_{l=1}^{L}\kappa_l E_l \equiv \delta_{\mathrm{network}}.
\label{eq:network_bound}
\end{equation}
The $\kappa_l$ are difficult to bound tightly in a real transformer, so Eq.~\ref{eq:network_bound} is a sufficient analytical statement rather than a practical runtime gate.

\begin{proposition}[Greedy decision stability]
Let $\ell^{\mathrm{full}}$ be the dense logits and let
\[
\gamma = \ell^{\mathrm{full}}_{(1)}-\ell^{\mathrm{full}}_{(2)}
\]
be the dense top-1 margin. If $\|\ell^{\mathrm{sparse}}-\ell^{\mathrm{full}}\|_\infty\le\delta$ and $2\delta<\gamma$, then sparse and dense attention have the same greedy argmax.
\end{proposition}

\begin{proof}
For the dense winner $v^*$ and any alternative $u$, the dense gap is at least $\gamma$. Each logit can move by at most $\delta$, so the sparse gap is at least $\gamma-2\delta>0$.
\end{proof}

For autonomous generation, this condition has to keep holding on the states actually produced by the sparse model. Matching one teacher-forced next-token decision does not guarantee a matching trajectory: once one token changes, the models condition on different prefixes and their trajectories bifurcate. Conversely, failing the infinity-norm condition does not imply the argmax changes; it is sufficient, not necessary.

Sparse and dense outputs are also not bit-identical because removing nonzero softmax terms changes the partition function. This corrects the bit-exact interpretation in our earlier preprint (arXiv:2602.06317). The relevant question here is greedy decision stability, not IEEE-754 equality.

\section{Selector and Evaluation Setup}
\subsection{Mean-Pooled Block Selection}
For query position $i$, the selector retains three pieces of history: position 0, the most recent $W=64$ tokens, and $r=2$ distant blocks of size $M=16$. Distant blocks are ranked by
\begin{equation}
s_b = \frac{q_i^\top \mu_b}{\sqrt d},\qquad
\mu_b = \frac{1}{M}\sum_{j\in B_b}k_j,
\end{equation}
using post-RoPE queries and keys where RoPE is present. The nominal support is
\begin{equation}
S = 1+W+rM = 97.
\end{equation}
Mean pooling is deliberately simple and is not claimed as the novelty of the paper; related page- and query-aware selection methods include QUEST, UNIQUE, TokenSelect, RetrievalAttention, and others \cite{tang2024quest,deng2025unique,wu2024tokenselect,liu2024retrievalattention}.

\paragraph{Budget-matched selector variants.}
We define four policies under a common cap on the union of distant positions read per KV head. \texttt{shared} ranks blocks using query and centroid averages shared across heads; \texttt{kv\_group} uses the mean query of each query-head group with the corresponding KV-head centroids; \texttt{head\_union} scores blocks for each query head and assigns each selected block to its best-scoring head, subject to the same per-KV-head cap; and \texttt{rerank} first forms a centroid shortlist, then reranks those blocks with token-level log-sum-exp scores. The \texttt{per\_head} reference selects independently for every query head and can exceed the common KV-read cap after taking the union within a GQA group. The clean survival sweep evaluates \texttt{kv\_group}; the other variants are specified comparators, not results of that sweep. Unique distant positions per KV head are recorded in addition to nominal per-head support.

At reuse interval $R$, distant blocks are selected at one-based sparse-decision indices $1,1+R,1+2R,\ldots$ and reused between scans while the local window continues to move. This separates support quality from temporal reuse: if $R=1$ already fails, reuse cannot be the only cause.

\subsection{Models and Tests}
The primary paired sweep uses \texttt{Qwen/Qwen2-0.5B}, a 24-layer RoPE model with 14 query heads and 2 KV heads (7:1 GQA), at contexts of 2K, 8K, and 16K. Five distinct WikiText-2 prefixes per context are paired across support sizes $S\in\{97,193,385,769\}$ with fresh selection ($R=1$) and a 128-token generation horizon. We report autonomous survival, teacher-forced token agreement, and perplexity change on the dense-greedy continuation. The first-token retrieval grid and H2O-style comparison below are earlier, separate 97-position probes; their outcomes are not pooled with the paired support sweep.

For each survival run, we first obtain a dense greedy continuation from the common prompt. The initial dense-prefill prediction is shared, so teacher-forced agreement and loss compare only later reference tokens, each evaluated from its preceding dense reference token. Thus $\Delta$PPL measures how the sparse next-token distribution scores the dense model's own continuation when the context is held to that path; it does not measure the quality of the sparse model's free-running output. It is not conventional held-out corpus perplexity. First divergence is indexed after the shared initial token, and survival is reported for the measured generation horizon.

Earlier GPT-2 fidelity tables are excluded because the archived prototype reconstructs part of the omitted attention tail and misaligns the teacher-forced cache. Those runs therefore do not measure pure sparse decoding and are not used as evidence here.

\subsection{Timing Scope}
The systems measurements are intentionally separate from the quality measurements. We report (i) a candidate-given full-sequence Triton kernel, where selection is excluded, and (ii) isolated single-query attention where fresh scan timing includes centroid scoring, top-$k$, gather, and sparse SDPA. Reuse timing uses a fixed pre-gathered sparse buffer. The reuse numbers are therefore component-based amortized estimates, not measured end-to-end rolling decode.

\section{Results}
\subsection{Autoregressive Decode Survival}
The main result is that local agreement substantially overstates autonomous stability. We vary the per-head support at three context lengths under fresh selection ($R=1$), recording the first sparse decision that differs from dense decoding, teacher-forced next-token agreement, and perplexity change on the dense-greedy continuation.

\begin{table}[ht]
\centering
\small
\renewcommand{\arraystretch}{1.15}
\resizebox{\linewidth}{!}{%
\begin{tabular}{|c|c|c|c|c|c|}
\hline
\textbf{Context} & \textbf{Support} & \textbf{Median $\Delta$PPL [min,max] (\%)} & \textbf{Mean TF [min,max] (\%)} & \textbf{Median Tdiv [range]} & \textbf{Survival at 128} \\
\hline
2K & 97  & +12.5 [+5.1,+69.4] & 88.5 [81.9,93.7] & 5 [1,14]  & 0/5 \\
2K & 193 & +2.6 [$-$0.8,+26.7]  & 92.5 [87.4,96.1] & 10 [7,14] & 0/5 \\
2K & 385 & +0.2 [$-$2.8,+11.6]  & 96.1 [93.7,98.4] & 21 [4,25] & 0/5 \\
2K & 769 & $-$0.3 [$-$1.8,+3.7]  & 97.6 [95.3,99.2] & 22 [14,43] & 0/5 \\
\hline
8K & 97  & +13.3 [$-$2.8,+71.7] & 89.8 [78.0,96.1] & 3 [1,9]  & 0/5 \\
8K & 193 & +4.5 [$-$2.9,+34.5]  & 91.5 [85.0,95.3] & 3 [1,9]  & 0/5 \\
8K & 385 & +1.6 [$-$1.9,+22.8]  & 93.6 [87.4,96.9] & 4 [1,17] & 0/5 \\
8K & 769 & $-$0.1 [$-$1.1,+15.1]  & 95.1 [89.0,97.6] & 11 [3,32] & 0/5 \\
\hline
16K & 97  & +40.8 [$-$1.3,+9877.0] & 62.5 [9.4,93.7] & 2 [1,6]  & 0/5 \\
16K & 193 & +25.7 [$-$6.7,+2688.4] & 70.1 [26.0,96.9] & 4 [2,14] & 0/5 \\
16K & 385 & +9.5 [$-$6.8,+497.5]  & 75.6 [40.9,96.9] & 4 [2,15] & 0/5 \\
16K & 769 & +4.6 [$-$5.1,+152.0]  & 87.7 [66.1,96.9] & 4 [3,15] & 0/5 \\
\hline
\end{tabular}%
}
\caption{Fresh-selection ($R=1$) support sweep on Qwen2-0.5B, with five paired WikiText-2 prefixes per context and 128 generated tokens per run. Brackets show the prompt-level minimum and maximum for $\Delta$PPL and teacher-forced (TF) match; medians can therefore be read alongside prompt-specific tails. $\Delta$PPL is computed on the dense-greedy continuation. Negative values indicate lower mean NLL on this sampled dense path, not a general quality improvement. Tdiv is the one-based sparse-decision index of the first exact token mismatch after the shared dense-prefill token.}
\label{tab:decode_survival}
\end{table}

The primary finding is that no run preserves exact greedy identity through 128 tokens; this does not mean every run suffers a comparable quality loss. For $S\ge193$, seven of nine cells have median $\Delta$PPL within 5\% of dense, but their prompt ranges matter. At 16K, prompt 0 has $\Delta$PPL of +927\%, +540\%, +321\%, and +152\% as support grows from 97 to 769; prompt 4 has +9877\%, +2688\%, +497\%, then +4.6\%. The min--max ranges in Table~\ref{tab:decode_survival} expose these tails alongside the medians.

The paired observations also show a candidate warning regime: all seven runs with TF match below 70\% have $\Delta$PPL above +100\%, and all seven come from prompts 0 and 4 at 16K. By contrast, TF match at or above 93\% has $\Delta$PPL ranging from $-$6.8\% to +11.6\%, not a tight $\pm6\%$ band. Because all low-TF observations share both the 16K context and these two prompt continuations, context and prompt identity are confounded with TF; this sample cannot distinguish a TF-associated warning from prompt- or context-specific effects. TF match and NLL also come from the same logits and targets. We therefore treat this as a descriptive association, not evidence of a universal or causal threshold, and as a falsifiable pattern for a broader prompt and model study.

Tdiv measures a different property: exact greedy-path identity. A single low-margin argmax flip ends survival even when average continuation NLL remains close, so Tdiv should not be read as a graded quality score. Conversely, a low TF match can accompany catastrophic loss increases, as the 16K tails demonstrate. These measurements characterize one model, selector, and prompt sample rather than a universal support law.

\subsection{Long-Context Retrieval}
As a separate earlier probe, we test first-token retrieval across three context lengths, five insertion depths, and three distinct factual keys using the 97-position protocol. These results are not a matched comparison with the layer-local support sweep above. Table~\ref{tab:retrieval_summary} aggregates the 15 trials at each context length; the full depth-by-depth grid is included in the appendix.

\begin{table}[ht]
\centering
\renewcommand{\arraystretch}{1.15}
\begin{tabular}{|c|c|c|c|c|}
\hline
\textbf{Context} & \textbf{Dense} & \textbf{$R=1$} & \textbf{$R=4$} & \textbf{$R=16$} \\
\hline
2K & 9/15 (60.0\%) & 4/15 (26.7\%) & 2/15 (13.3\%) & 2/15 (13.3\%) \\
4K & 9/15 (60.0\%) & 2/15 (13.3\%) & 1/15 (6.7\%) & 1/15 (6.7\%) \\
8K & 9/15 (60.0\%) & 2/15 (13.3\%) & 1/15 (6.7\%) & 0/15 (0.0\%) \\
\hline
\textbf{Total} & \textbf{27/45 (60.0\%)} & \textbf{8/45 (17.8\%)} & \textbf{4/45 (8.9\%)} & \textbf{3/45 (6.7\%)} \\
\hline
\end{tabular}
\caption{First-token retrieval on Qwen2-0.5B with a 97-position support. Each context aggregates five insertion depths and three facts. Dense attention succeeds on 27/45 prompts, while fresh sparse selection succeeds on 8/45. Reuse reduces that further. These are first-token retrieval results, not full-answer accuracy.}
\label{tab:retrieval_summary}
\end{table}

This result matters because it puts the survival failure in context. The selector is not merely accumulating tiny harmless errors over hundreds of steps; it is already missing functionally important context often enough that fresh sparse inference underperforms dense attention sharply on a direct retrieval test.

\subsection{What Mean-Pooled Retrieval Recovers}
The selector can still look strong under attention-mass diagnostics. On a separate 2,042-token probe, $M=16$ mean-pooled blocks recover nearly the same \emph{total attention mass} as an exhaustive dense block oracle in deeper layers, even though end-to-end retrieval and autonomous decoding remain poor.

\begin{table}[ht]
\centering
\small
\renewcommand{\arraystretch}{1.15}
\resizebox{\linewidth}{!}{%
\begin{tabular}{|c|c|c|c|c|c|}
\hline
\textbf{Layer} & \textbf{Base Mass} & \textbf{Oracle Total} & \textbf{Mean-Pool Total} & \begin{tabular}[c]{@{}c@{}}\textbf{Oracle}\\\textbf{Distant Recall}\end{tabular} & \begin{tabular}[c]{@{}c@{}}\textbf{Mean-Pool}\\\textbf{Distant Recall}\end{tabular} \\
\hline
12 & 79.59\% & 92.52\% & 92.30\% & 60.28\% & 53.75\% \\
18 & 84.99\% & 94.63\% & 94.63\% & 47.13\% & 46.51\% \\
22 & 69.16\% & 89.52\% & 89.47\% & 56.07\% & 55.70\% \\
\hline
\end{tabular}%
}
\caption{Mass-recovery diagnostic for $M=16$, $r=2$, and total support 97 on one Qwen2-0.5B probe. Mean pooling can match the oracle's total captured mass closely in deeper layers while still failing badly on downstream retrieval and autonomous decode. Attention mass alone is therefore not a sufficient quality metric.}
\label{tab:mass_recovery}
\end{table}

This is an important distinction for the Condensation Principle. Concentration makes sparse inference plausible; it does not certify a sparse decision. Which values are omitted and how their error propagates matter just as much as the amount of probability mass retained.

An exploratory block-size sweep on this probe reported dynamic recall of 31.0--41.6\% at $M=32$, compared with 46.5--55.7\% at $M=16$, and 17.9--31.2\% at $M=64$. The $M=64$ setting uses 129 active positions rather than 97, so it is not a matched-budget comparison; the single-probe trend is consistent with within-block dilution, not evidence of a universal block-size threshold.

\subsection{H2O-Style Comparison}
We also compare current-query selection with a head-wise cumulative-heavy-hitter policy modeled after H2O \cite{zhang2023h2o}. The comparison uses non-repeating WikiText-2 text, causal accumulation, and matched per-head attention-access budgets. It is a simulation of the selection rule, not the official H2O serving implementation.

\begin{table}[ht]
\centering
\small
\renewcommand{\arraystretch}{1.15}
\resizebox{0.96\linewidth}{!}{%
\begin{tabular}{|l|c|c|c|}
\hline
\textbf{Method} & \textbf{Budget} & \textbf{First-Token Retrieval} & \textbf{Sampled PPL} \\
\hline
Dense & Full & 17/30 (56.7\%) & 12.02 \\
H2O-style cumulative & 97/head & 0/30 (0.0\%) & 26.48 \\
Current-query mean-pool & 97/head & 9/30 (30.0\%) & 21.53 \\
H2O-style cumulative & Adaptive 20\% & 10/30 (33.3\%) & 21.26 \\
Current-query mean-pool & Adaptive 20\% & --- & 12.91 \\
\hline
\end{tabular}%
}
\caption{Two small supporting probes on Qwen2-0.5B. Retrieval uses 30 constructed needle configurations at 1K--2K; sampled PPL uses 32 positions from one 1,024-token WikiText-2 stream. The metrics come from different probes and should not be treated as a single benchmark. At the matched 97-position budget, current-query selection is better than the cumulative simulation but remains far from dense.}
\label{tab:h2o_compact}
\end{table}

The baseline comparison therefore supports a limited claim: current-query retrieval can be better than backward-looking cumulative selection at the same small budget, especially for cold context. It does not rescue the 97-position operating point; both sparse methods remain substantially worse than dense attention in these tests.

\subsection{Attention Cost}
The quality results above are poor at the tested budget, but the computational incentive is real. We report two different timing workloads and keep them separate.

\paragraph{Candidate-given full-sequence kernel.}
For $N$ queries each attending to a supplied set of 96 keys, the Triton kernel scales as $O(NSd)$ rather than dense $O(N^2d)$. Candidate discovery is excluded.

\begin{table}[ht]
\centering
\small
\renewcommand{\arraystretch}{1.1}
\begin{tabular}{|r|r|r|r|}
\hline
\textbf{Sequence} & \textbf{SDPA} & \textbf{Sparse} & \textbf{Speedup} \\
\hline
8,192 & 3.27 ms & 0.19 ms & 16.8$\times$ \\
32,768 & 41.41 ms & 1.05 ms & 39.3$\times$ \\
65,536 & 156.25 ms & 2.78 ms & 56.1$\times$ \\
131,072 & 627.58 ms & 3.94 ms & 159.3$\times$ \\
\hline
\end{tabular}
\caption{Candidate-given full-sequence attention on GPT-2-sized tensors ($B=1,H=12,d=64$). Candidate discovery and non-attention model components are excluded. The frozen benchmark uses 96 supplied slots per query, rather than the 97-position policy used in the Qwen experiments.}
\label{tab:kernel_compact}
\end{table}

\paragraph{Single-query attention.}
At 131K KV length, the fresh scan includes centroid scoring, top-$k$, real gathers, and sparse SDPA. Reuse uses a fixed pre-gathered buffer; amortized reuse figures are estimates from component medians.

\begin{table}[ht]
\centering
\small
\renewcommand{\arraystretch}{1.15}
\resizebox{\linewidth}{!}{%
\begin{tabular}{|l|c|c|c|c|c|}
\hline
\textbf{Configuration} & \textbf{Dense FA-2} & \textbf{Fresh Scan} & \textbf{Fresh Speedup} & \textbf{Reuse Component} & \textbf{Amortized Estimate} \\
\hline
GPT-2-sized, $R=4$ & 2.137 ms & 0.693 ms & 3.1$\times$ & 0.013 ms & 0.183 ms (11.7$\times$) \\
70B-scale GQA, $R=16$ & 3.064 ms & 2.081 ms & 1.5$\times$ & 0.016 ms & 0.145 ms (21.1$\times$) \\
\hline
\end{tabular}%
}
\caption{Isolated 131K-token single-query attention on an RTX 4090 Laptop GPU under Linux/WSL2. The reuse figures are component-based estimates, not measured rolling decode. KV appends, rolling-window updates, centroid maintenance, host transfers, projections, MLPs, and vocabulary logits are excluded.}
\label{tab:decode_timing_compact}
\end{table}

These numbers show that long-context sparse attention can reduce the amount of data touched by the attention operator. They do not show matched-quality model acceleration, because the 97-position Qwen operating point tested above is not quality-preserving and the benchmarked kernel is not independently reproducible from the public artifact.

\section{Discussion}
The experiments change the interpretation of the original idea in several ways.

\paragraph{Attention concentration is only the starting point.}
The mass-recovery probe can look encouraging while retrieval and autonomous decode fail badly. The relevant object is not just omitted probability mass but the omitted value contribution after downstream propagation. A selector can retain more than 90\% of attention mass and still miss the positions that matter for a particular decision.

\paragraph{Exact-match survival and distributional quality are separate axes.}
For $S\ge193$, seven of nine context-support cells have median continuation $\Delta$PPL within 5\% of dense, while the min--max ranges show severe outliers at 16K. All seven runs with TF match below 70\% have $\Delta$PPL above +100\%, but they come from only two prefixes. The apparent transition is worth testing; it is not yet a general threshold. Teacher-forced PPL measures average next-token likelihood on a fixed dense continuation, not semantic task quality or the sparse model's own free-running continuation. Tdiv remains relevant to exact decision stability, but is not a proxy for quality-loss magnitude.

\paragraph{Fresh selection does not make the tested support range stable.}
In the clean $R=1$ sweep, all supports from 97 through 769 diverge within 43 sparse decisions. Earlier $R=4$ and $R=16$ results at support 97 used a separate shared-selection harness and are not pooled with this layer-local sweep. Reuse at larger supports remains unmeasured, so these data do not establish whether reuse helps once retrieval quality improves.

\paragraph{The timing result and the quality result are currently disconnected.}
The attention operator can be much cheaper when it reads a small support. The present selector does not preserve Qwen2-0.5B well enough at $S=97$ to call those savings matched-quality acceleration. Establishing that claim would require a support/selector regime that survives autonomous decoding, followed by timing on that exact regime.

\subsection{Limitations}
The empirical boundary is narrow. The clean support sweep uses one small model and five natural prefixes at each of three context lengths; its 60 runs are paired measurements over 15 prefixes and four supports, not 60 independent prompts. The earlier first-token retrieval grid and reuse results use a separate selector protocol and are reported as supporting evidence, not direct matched comparisons with the clean sweep. Mistral-7B remains uncharacterized beyond mechanics smoke tests. None of these measurements establishes a population failure rate or a universal sparse-attention limit.

The selector itself is intentionally simple. Mean pooling can dilute isolated high-scoring keys, and a fixed 97-position support becomes proportionally smaller as context grows. We have not yet separated four plausible causes of failure: coarse retrieval error, insufficient support size, harmful omitted value directions despite high retained mass, and recursive cache drift. The fact that $R=1$ fails early only rules out reuse as the sole explanation.

The systems results are also limited. The public repository does not reproduce the proprietary optimized kernel used for the headline timing measurements. Candidate-given full-sequence timing excludes discovery, and reuse timings are composed from separately measured components rather than a rolling decode loop. Host-backed KV transfers and whole-model TPOT are not measured. These should be treated as operator-level observations until a reproducible matched-quality implementation exists.

\section{Related Work}
Sparse and efficient attention spans several distinct approaches. Fixed sparse patterns such as Sparse Transformers, Longformer, and BigBird reduce pairwise interactions through predetermined structures \cite{child2019sparsetransformers,beltagy2020longformer,zaheer2020bigbird}. Exact systems such as FlashAttention, Flash-Decoding, FlashInfer, and PagedAttention reduce memory movement without changing the dense mathematical result \cite{dao2022flashattention,dao2023flashdecoding,ye2025flashinfer,kwon2023pagedattention}. StreamingLLM, InfLLM, and DuoAttention exploit sinks, local windows, or head specialization for long-context serving \cite{xiao2023streamingllm,xiao2024infllm,xiao2024duoattention}.

Closer to this paper, query-aware KV selection retrieves a small subset conditioned on the current query. QUEST uses page-level query-aware bounds \cite{tang2024quest}; TokenSelect performs per-head dynamic token selection \cite{wu2024tokenselect}; RetrievalAttention uses vector retrieval over historical KV \cite{liu2024retrievalattention}; UNIQUE uses page statistics for training-free top-$k$ sparse attention \cite{deng2025unique}; Tactic and related methods adapt sparse budgets to the observed attention distribution \cite{zhu2025tactic}. H2O instead retains recent tokens and cumulative historical heavy hitters \cite{zhang2023h2o}. Trainable approaches such as SeerAttention and Native Sparse Attention learn sparse routing or structure directly \cite{gao2024seerattention,yuan2025native}.

This paper does not claim novelty for attention concentration, query-aware selection, or mean-pooled blocks. Its focus is narrower: connect omitted attention to a greedy-margin condition, then test whether a concrete low-budget selector actually preserves autonomous decoding. The negative survival result is central because it shows that mass concentration and strong teacher-forced agreement can coexist with immediate recursive failure.

\section{Conclusion}
The Condensation Principle is a conditional statement, not a blanket claim that trained attention can always be sparsified. If attention concentrates on a small query-dependent support and the omitted contribution stays inside the downstream decision margin, sparse attention can preserve the dense greedy decision. The experiments here show how easily the second half of that condition can fail.

For Qwen2-0.5B, the primary result is 0/60 exact trajectories surviving 128 tokens. Distributional quality is a distinct finding: seven of nine $S\ge193$ cells have median continuation $\Delta$PPL within 5\%, while the 16K prompt-level range reaches +9877\% at $S=97$ and +152\% at $S=769$. All seven runs below 70\% teacher-forced match exceed +100\% $\Delta$PPL, concentrated in two prefixes; this suggests a warning regime, not a validated threshold. The measured operating envelope is model- and prompt-specific, and the separate retrieval and timing probes do not establish matched-quality acceleration.

The next question is which retrieval policy reaches the best quality-cost operating point under equal KV-read budgets, and whether any such point preserves autonomous decoding. The current comparisons do not identify a globally best selector.

\section*{Code and Data Availability}
Validation scripts and reference implementations are available at \url{https://github.com/JorgeLRW/condensate-theorem}. The public code does not reproduce the proprietary optimized Triton kernel used for the headline timing results. Timing logs, run metadata, and quality tables should be archived with the submission so the reported operator-level measurements can be audited independently.

\section*{Acknowledgments}
This manuscript substantially revises arXiv:2602.06317, which incorrectly described sparse attention as bit-exact and lossless at the floating-point output level. The current version replaces that claim with a margin-based view and reports cases where the tested sparse selector fails to preserve autonomous decoding. We thank Halil Talha Ertan for correspondence that helped expose the earlier bit-exact error and sharpen the candidate-selection analysis.

\appendix
\section{Full Long-Context Retrieval Grid}
The main text reports context-level aggregates. For completeness, Table~\ref{tab:retrieval_full} gives the original depth-by-depth first-token results.

\begin{table}[ht]
\centering
\scriptsize
\renewcommand{\arraystretch}{1.05}
\resizebox{\linewidth}{!}{%
\begin{tabular}{|c|c|c|c|c|c|}
\hline
\textbf{Context} & \textbf{Depth} & \textbf{Dense} & \textbf{$R=1$} & \textbf{$R=4$} & \textbf{$R=16$} \\
\hline
\multirow{5}{*}{2K}
&10\%&2/3&1/3&0/3&0/3\\
&25\%&1/3&0/3&0/3&0/3\\
&50\%&2/3&2/3&2/3&2/3\\
&75\%&2/3&1/3&0/3&0/3\\
&90\%&2/3&0/3&0/3&0/3\\
\hline
\multirow{5}{*}{4K}
&10\%&2/3&0/3&0/3&0/3\\
&25\%&2/3&0/3&0/3&0/3\\
&50\%&2/3&1/3&1/3&1/3\\
&75\%&2/3&1/3&0/3&0/3\\
&90\%&1/3&0/3&0/3&0/3\\
\hline
\multirow{5}{*}{8K}
&10\%&2/3&1/3&0/3&0/3\\
&25\%&2/3&0/3&0/3&0/3\\
&50\%&2/3&0/3&0/3&0/3\\
&75\%&1/3&0/3&0/3&0/3\\
&90\%&2/3&1/3&1/3&0/3\\
\hline
\end{tabular}%
}
\caption{Depth-by-depth version of the Qwen2-0.5B first-token retrieval grid summarized in Table~\ref{tab:retrieval_summary}.}
\label{tab:retrieval_full}
\end{table}

\section{Notes on Timing Interpretation}
The candidate-given kernel, fresh-scan path, and reuse estimate are different workloads. The full-sequence kernel assumes candidate indices are already available. Fresh single-query scan measures centroid scoring, selection, gather, and sparse attention with centroids precomputed outside the timed region. Reuse timing measures sparse attention on a fixed gathered buffer and therefore omits cache maintenance and refreshed gathers. None of these measurements is whole-model TPOT.

At 131K tokens, the GPT-2-sized dense KV payload is 402.7 MB and the measured FlashAttention-2 time is 2.137 ms, corresponding to a nominal payload/time ratio of 188.4 GB/s. The synthetic 70B-scale GQA payload is 536.9 MB with a measured dense time of 3.064 ms, or 175.2 GB/s by the same nominal calculation. These are not hardware-counter bandwidth measurements.


\begin{thebibliography}{99}
\sloppy
\hbadness=10000
\emergencystretch=3em

\bibitem{beltagy2020longformer}
Beltagy, I., Peters, M.~E., and Cohan, A. (2020).
Longformer: The Long-Document Transformer.
\textit{arXiv preprint arXiv:2004.05150}.

\bibitem{child2019sparsetransformers}
Child, R., Gray, S., Radford, A., and Sutskever, I. (2019).
Generating Long Sequences with Sparse Transformers.
\textit{arXiv preprint arXiv:1904.10509}.

\bibitem{dao2022flashattention}
Dao, T., Fu, D.~Y., Ermon, S., Rudra, A., and R{\'e}, C. (2022).
FlashAttention: Fast and Memory-Efficient Exact Attention with IO-Awareness.
In \textit{Advances in Neural Information Processing Systems (NeurIPS 2022)}, 35:16344--16359.

\bibitem{dao2023flashdecoding}
Dao, T., Haziza, D., Massa, F., and Chen, B. (2023).
Flash-Decoding for Long-Context Inference.
\textit{Stanford CRFM and Together AI Blog}.

\bibitem{deng2025unique}
Deng, K., Ling, S., Fan, R., and Li, J. (2025).
UNIQUE: Universal Top-$k$ Sparse Attention for Training-free Inference and Sparsity-aware Training.
\textit{arXiv preprint arXiv:2605.27740}.

\bibitem{gao2024seerattention}
Gao, Y., Xing, Z., Du, X., et al. (2024).
SeerAttention: Learning Intrinsic Sparse Attention in Your LLMs.
\textit{arXiv preprint arXiv:2410.13276}.

\bibitem{kwon2023pagedattention}
Kwon, W., Li, Z., Zhuang, S., Sheng, Y., Zheng, L., Yu, C.~H., Gonzalez, J.~E., Zhang, H., and Stoica, I. (2023).
Efficient Memory Management for Large Language Model Serving with PagedAttention.
In \textit{Proceedings of the 29th ACM Symposium on Operating Systems Principles (SOSP 2023)}, pp.~611--626.

\bibitem{liu2024retrievalattention}
Liu, D., Chen, M., Lu, B., Jiang, H., Han, Z., Zhang, Q., Chen, Q., Zhang, C., Ding, B., Zhang, K., Chen, C., Yang, F., Yang, Y., and Qiu, L. (2024).
RetrievalAttention: Accelerating Long-Context LLM Decoding via Vector Retrieval.
In \textit{Advances in Neural Information Processing Systems (NeurIPS 2024)}.

\bibitem{tang2024quest}
Tang, J., Zhao, Y., Zhu, K., Xiao, G., Kasikci, B., and Han, S. (2024).
QUEST: Query-Aware Sparsity for Efficient Long-Context LLM Inference.
In \textit{Proceedings of the 41st International Conference on Machine Learning (ICML 2024)}, PMLR 235:47372--47395.

\bibitem{wu2024tokenselect}
Wu, W., Pan, Z., Wang, C., Chen, L., Bai, Y., Wang, T., Fu, K., Wang, Z., and Xiong, H. (2024).
TokenSelect: Efficient Long-Context Inference and Length Extrapolation for LLMs via Dynamic Token-Level KV Cache Selection.
In \textit{Findings of the Association for Computational Linguistics: EMNLP 2024}, arXiv:2411.02886.

\bibitem{xiao2023streamingllm}
Xiao, G., Tian, Y., Chen, B., Han, S., and Lewis, M. (2023).
Efficient Streaming Language Models with Attention Sinks.
In \textit{International Conference on Learning Representations (ICLR 2024)}, arXiv:2309.17453.

\bibitem{xiao2024infllm}
Xiao, C., Zhang, P., Han, X., Xiao, G., Lin, Y., Zhang, Z., Liu, Z., and Sun, M. (2024).
InfLLM: Training-Free Long-Context Extrapolation for LLMs with Memory Units.
In \textit{Advances in Neural Information Processing Systems (NeurIPS 2024)}, arXiv:2402.04617.

\bibitem{xiao2024duoattention}
Xiao, G., Tang, J., Han, S., et al. (2024).
DuoAttention: Efficient Long-Context LLM Inference with Retrieval and Streaming Heads.
\textit{arXiv preprint arXiv:2410.10819}.

\bibitem{ye2025flashinfer}
Ye, Z., Chen, L., Lai, R., et al. (2025).
FlashInfer: Efficient and Customizable Attention Engine for LLM Inference Serving.
\textit{arXiv preprint arXiv:2501.01005}.

\bibitem{yuan2025native}
Yuan, J., Gao, Y., Xing, Z., et al. (2025).
Native Sparse Attention: Hardware-Aligned and Natively Trainable Sparse Attention.
\textit{arXiv preprint arXiv:2502.11089}.

\bibitem{zaheer2020bigbird}
Zaheer, M., Guruganesh, G., Dubey, K.~A., Ainslie, J., Alberti, C., Ontanon, S., Pham, P., Ravula, A., Wang, Q., Yang, L., and Ahmed, A. (2020).
Big Bird: Transformers for Longer Sequences.
In \textit{Advances in Neural Information Processing Systems (NeurIPS 2020)}, 33:17283--17297.

\bibitem{zhang2023h2o}
Zhang, Z., Sheng, Y., Zhou, T., Chen, T., Zheng, L., Cai, R., Song, Z., Tian, Y., R{\'e}, C., Barrett, C., Wang, Z., and Chen, B. (2023).
H2O: Heavy-Hitter Oracle for Efficient Generative Inference of Large Language Models.
In \textit{Advances in Neural Information Processing Systems (NeurIPS 2023)}, 36:34661--34730.

\bibitem{zhu2025tactic}
Zhu, K., Tang, T., Xu, Q., Gu, Y., Zeng, Z., Kadekodi, R., Zhao, L., Li, A., Krishnamurthy, A., and Kasikci, B. (2025).
Tactic: Adaptive Sparse Attention with Clustering and Distribution Fitting for Long-Context LLMs.
\textit{arXiv preprint arXiv:2502.12216}.

\end{thebibliography}
\end{document}